\documentclass{article}

\PassOptionsToPackage{numbers, compress}{natbib}
\usepackage[preprint]{neurips_2026}

\usepackage[utf8]{inputenc} % allow utf-8 input
\usepackage[T1]{fontenc}    % use 8-bit T1 fonts
\usepackage{hyperref}       % hyperlinks
\usepackage{url}            % simple URL typesetting
\usepackage{booktabs}       % professional-quality tables
\usepackage{amsfonts}       % blackboard math symbols
\usepackage{nicefrac}       % compact symbols for 1/2, etc.
\usepackage{microtype}      % microtypography
\usepackage{xcolor}         % colors

\usepackage{amsmath}
\usepackage{amssymb}
\usepackage{mathtools}
\usepackage{amsthm}
\usepackage{bm}
\usepackage{siunitx}
\usepackage{algorithm}
\usepackage{algorithmic}
\usepackage{subcaption}

\title{
TRIE: An Evaluation Framework for Stochastic PDE Surrogates
}

\author{
  Bharat Srikishan\textsuperscript{1}\thanks{Corresponding Authors: \texttt{bsrikish@stevens.edu, cyoung@lanl.gov}} \\
  \And
  Javier E. Santos\textsuperscript{2} \\
  \And
  Nikhil Muralidhar\textsuperscript{1} \\
  \And
  Charles D. Young\textsuperscript{2}$^*$ \\
  \AND
  \textsuperscript{1} \textnormal{Stevens Institute of Technology}
  \textsuperscript{2} \textnormal{Los Alamos National Laboratory}
}

\begin{document}

\maketitle

\begin{abstract}

  Many scientific systems exhibit uncertainty from stochastic forcing, unresolved degrees of freedom, or imperfect observations, making reliable surrogate forecasting fundamentally distributional rather than pointwise. For such systems, deterministic neural surrogates fail to capture statistical measures and forecast uncertainty.  We introduce TRIE, an evaluation framework for stochastic PDE surrogates that asks whether models reproduce invariant measures, provide trustworthy predictive uncertainty, and scale to efficient probabilistic generation. We demonstrate TRIE on two stationary chaotic spatially extended SPDEs, stochastic Kuramoto--Sivashinsky and stochastic Kolmogorov flow, across 11 parameter values. Our evaluation shows that standard pointwise-trained neural surrogates can produce plausible short rollouts while failing to match long-time statistical structure. Approximate uncertainty methods such as Monte Carlo dropout and heteroscedastic Gaussian likelihoods produce stochastic forecasts, but are often miscalibrated and overconfident under temporal and spatial uncertainty diagnostics. Across these criteria, generative models provide the most consistent performance, accurately capturing invariant measure statistics and achieving the lowest CRPS in all reported probabilistic settings. Finally, we show that latent generative models with automatic dimension discovery retain much of this statistical fidelity while reducing Kolmogorov inference time by roughly $12\times$. We release our code\footnote{https://github.com/scailab/TRIE-SPDE-Bench} and data to support reproducible evaluation of stochastic PDE forecasting models.
\end{abstract}

\section{Introduction}

Real world data and processes are inevitably stochastic, leading to uncertainty in measurements and predictions. In some cases, such as low Gaussian measurement noise, the effect is minimal, and deterministic models are reliable for some applications. Often, the observed state does not contain enough information about the stochastic process. Molecular systems experience seemingly random thermal fluctuations which can deterministically explained by fast atomistic degrees of freedom, but the latter cannot be tractably measured or computed. Similar effects common in physical, chemical, and biological systems \cite{horsthemke1984noise, van1992stochastic}, where unknown stochastic forcing couples to the dynamics. Far from equilibrium, stochastic and deterministic effects couple to drive diverse spatiotemporal patterns and transitions \cite{cross1993pattern}. These concepts extend to weather \cite{leutbecher2008ensemble}, finance \cite{voit2005statistical}, and power grids \cite{anghel2007stochastic}. 

Here we focus on stochastic partial differential equations (SPDEs) as models for spatially extended systems under stochastic forcing. In nonlinear parameter regimes, time and length scales are often coupled over orders of magnitude, requiring computationally expensive fine discretizations and specialized numerical solvers. Deterministic surrogates can significantly reduce computational expense while retaining short time accuracy \cite{pathak2018model}, and improvements have been introduced to improve stability \cite{linot2023stabilized, pedersen2025thermalizer}, multi-step rollouts \cite{bengio2015scheduled}, and physical constraints \cite{meng2025physics}. However, they are limited in that by construction, they are unable to produce forecast distributions. Approximate uncertainty methods such as Monte Carlo dropout \cite{gal2016dropout}, ensembles \cite{lakshminarayanan2017simple}, and Kalman filters \cite{hamilton2016ensemble}, provide uncertainty estimates, but may be miscalibrated or computationally expensive \cite{djupskaas2025unreliable,folgoc2021mc}. Recent probabilistic generative models, including diffusion-based forecasters and stochastic interpolants, offer a more direct route to distributional forecasting \cite{ruhling2023dyffusion,yu2024diffcast,price2025gencast,gao2024generative,rozet2025lost,albergo2023stochastic,chen2024probabilistic,albergo2025stochastic}.

We introduce TRustworthiness, Invariance, and Efficiency criteria or \textbf{TRIE}, an evaluation framework for stochastic PDE surrogates. TRIE asks whether a model: (i) reproduces long-time invariant measures, (ii) provides trustworthy predictive uncertainty, and (iii) scales to efficient probabilistic generation. We evaluate invariant measure fidelity using derivative-field joint densities and spectral content. We evaluate probabilistic trustworthiness using the continuous ranked probability score (CRPS) and spatial uncertainty diagnostics. Finally, we evaluate generative scaling through wall-clock inference time, since high-quality probabilistic samplers can be substantially slower than deterministic surrogates. To address inference cost, we study latent stochastic forecasting with implicit rank minimizing autoencoders \cite{jing2020implicit, zeng2024autoencoders}.

We demonstrate TRIE on one-dimensional stochastic Kuramoto--Sivashinsky and two-dimensional stochastic Kolmogorov for across 11 parameter values. Across deterministic, approximate probabilistic, flow-based, and stochastic interpolant surrogates, we find that pointwise-trained models often fail to reproduce invariant measures, and approximate uncertainty methods can be overconfident or miscalibrated. Stochastic interpolants provide the most consistent performance. Latent stochastic interpolants retain much of this statistical fidelity while reducing Kolmogorov inference time by roughly $12\times$.

Our main contributions are:
\begin{itemize}
    \item We introduce TRIE, an evaluation framework for SPDE surrogates based on invariant measure fidelity, probabilistic trustworthiness, and generative scaling.
    \item We show that invariant measure metrics and CRPS reveal failures hidden by pointwise evaluation, with distributional bridging surrogates achieving the most consistent performance.
    \item We apply automatic dimension discovery \cite{jing2020implicit, zeng2024autoencoders} to SPDEs for the first time to accelerate inference with minimal parameter tuning while maintaining TRIE reliability.
    \item We release simulation, training, inference, and evaluation \href{https://github.com/scailab/TRIE-SPDE-Bench}{code} for two sophisticated SPDE systems.
\end{itemize}

\begin{figure}[!t]
\centering
\includegraphics[width=\textwidth]{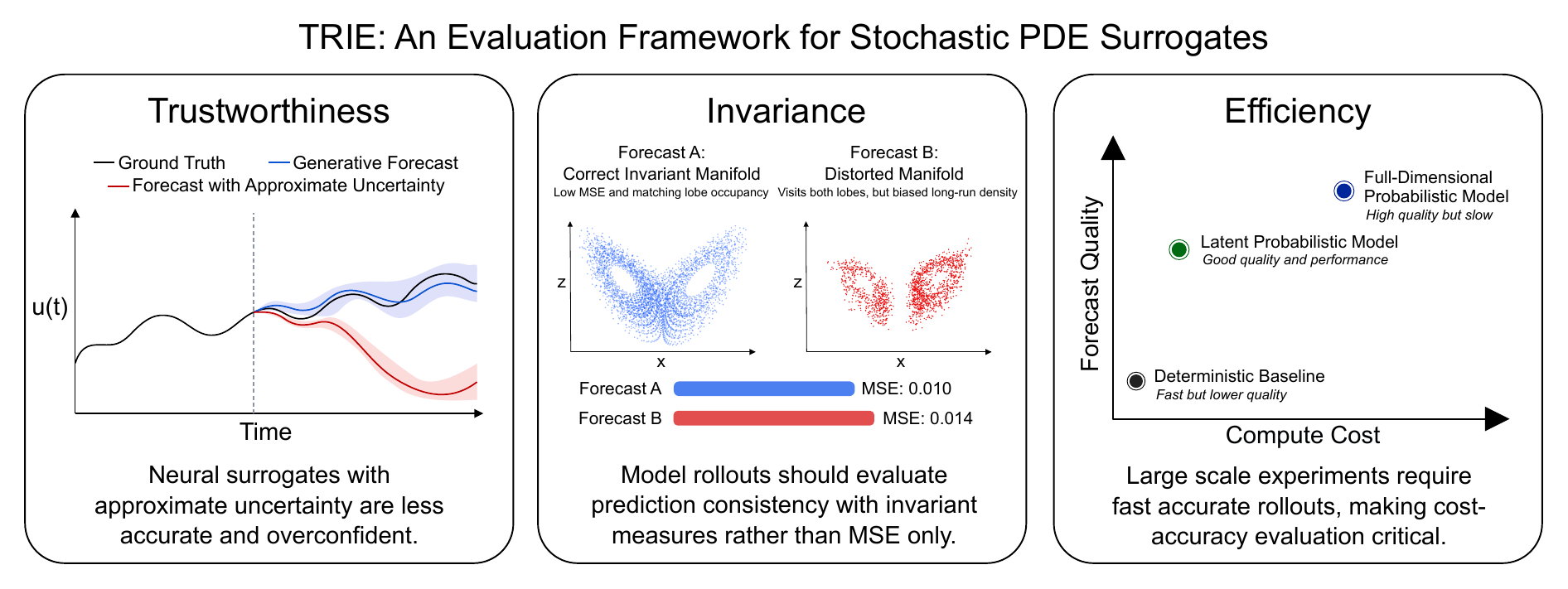}
\caption{\textbf{Trustworthiness}: Neural surrogates with approximate or absent uncertainty quantification are often overconfident when forecast distributions deviate from the true density. The continuous ranked probability score (CRPS) and spatial uncertainty metrics capture these often overlooked nuances. \textbf{Invariant Measures} are a reliable indicator to whether the surrogate has learned the true dynamical system or simply fit short-time dynamics. \textbf{Efficiency}: Evaluating surrogate solutions on their ability to navigate the cost-accuracy tradeoff with inference wall-clock time is paramount.}
\label{fig:overview}
\end{figure}

\section{Related Work}

\textbf{Benchmarks and learned stochastic dynamics.}
Neural PDE benchmarks typically emphasize short-horizon prediction error, rollout stability, or architecture comparisons. Recent work has extended this direction to stochastic systems. Neural SPDEs learn solution operators conditioned on both an initial state and a realization of the driving noise \cite{salvi2022neural}, while SPDEBench evaluates models for regular and singular SPDEs with emphasis on numerical schemes, noise sampling, renormalization, and comparisons with models such as FNO, NSPDE, and DLR-Net \cite{li2025spdebench}. TRIE is complementary: rather
than focusing primarily on numerical discretization or architecture-level comparisons, we evaluate whether stochastic forecasters reproduce invariant measures, provide calibrated and sharp predictive distributions, and generate samples efficiently. This distinction is important for chaotic stochastic systems, where pointwise error can obscure failures in long-time statistical structure or uncertainty calibration.

\textbf{Probabilistic forecasters and uncertainty quantification.}
Probabilistic generative models provide a natural way to model conditional forecast distributions rather than point estimates. Flow matching trains continuous normalizing flows by regressing vector fields along conditional probability paths \cite{lipman2023flow}, and stochastic interpolants unify flow-based and diffusion-based generative models through continuous-time bridges between probability distributions \cite{albergo2025stochastic}. We evaluate these methods as single-model conditional forecasters under a shared one-step training and autoregressive rollout protocol. Approximate uncertainty methods such as Monte Carlo dropout and heteroscedastic Gaussian likelihoods are computationally attractive, while deep ensembles are a strong alternative for calibrated uncertainty estimation \cite{lakshminarayanan2017simple}. With TRIE we quantitatively evaluate the calibration of these samples in conjunction with long time rollout stability and accuracy.

\textbf{Long-time statistics and rollout correction.}
For chaotic dissipative systems, long-time fidelity is often statistical rather than trajectory-wise: rollouts may diverge pointwise while still sampling the correct attractor. Several methods improve long-horizon behavior by modifying autoregressive inference with denoising refinements and projections\cite{lippe2023pde, pedersen2025thermalizer}. These methods are closely related to our invariant measure criteria, but they alter the forecasting procedure by adding correction steps during inference. In contrast, TRIE evaluates the raw one-step surrogate itself.

\textbf{Reduced order and latent-space generation.}
Iterative probabilistic samplers can be expensive at full spatial resolution, motivating reduced order approaches. Classical projection methods exploit low-dimensional structure in dissipative systems \cite{fletcher1984computational}, while modern autoencoder-based approaches learn nonlinear latent coordinates for dynamics and physics emulation \cite{koronaki2024nonlinear,zeng2024autoencoders}. Recent latent generative models show that sampling in compressed coordinates can substantially reduce inference cost \cite{rozet2025lost}. TRIE evaluates this cost by testing whether latent stochastic interpolants can retain invariant measure fidelity while reducing generation time. We consider implicit rank-minimizing autoencoders \cite{jing2020implicit}, which have shown strong performance in automatic latent size discovery for deterministic chaos \cite{zeng2024autoencoders}. We extend to stochastic systems on random attractors \cite{crauel1997random}, showing the same success in automatic dimension discovery.

\section{Methods}

\subsection{TRIE: TRustworthiness, Invariance, and Efficiency}

We consider stochastic partial differential equations (SPDEs) governing the evolution of a state a state $\bm{x}(t)$ of the form:
\begin{equation}
    \partial_t \bm{x} = f(\bm{x},t) + \eta(\bm{x},t), \quad t \in [0, T], \bm{x} \in \mathbb{R}^d
\end{equation}
where $f$ denotes the deterministic dynamics and $\eta$ denotes stochastic forcing. In this work we focus on state-independent noise processes $\eta(t)$, although the evaluation framework applies more broadly to state-dependent stochastic systems. Since the forcing realization is not observed at forecast time, a surrogate cannot track a single reference trajectory indefinitely. Instead, for chaotic and stochastic systems, long rollouts should reproduce the statistical structure of
the dynamics after transients have decayed.

We therefore evaluate SPDE surrogates using three complementary questions. First, \textbf{can neural surrogates accurately capture invariant measures?} This question is central because a surrogate may produce plausible short forecasts while converging to the wrong stochastic attractor or assigning incorrect probability mass. Second, \textbf{how trustworthy are surrogate predictions?} In scientific applications, forecasts are used to guide analysis, design, and downstream decisions. A useful model must therefore provide uncertainty estimates that are both calibrated and sharp, rather than merely producing diverse samples. Third, \textbf{can probabilistic surrogates be scaled for faster generation?} Generative forecasters can improve statistical fidelity, but their iterative sampling procedures can be expensive at full spatial resolution. Practical SPDE forecasting therefore requires measuring and improving inference-time cost in addition to accuracy. Below we define metrics and methodology to address these questions.

\subsection{Invariant Measures}

For dissipative stochastic systems, after an initial transient time $\tau$, trajectories concentrate near a random attractor \cite{crauel1997random, robinson2003infinite} and induce an invariant measure over state space. This measure describes the long-time distribution of states visited by the SPDE under the combined effects of deterministic dynamics and stochastic forcing. We evaluate whether autoregressive surrogate rollouts reproduce this distribution by comparing empirical invariant measure statistics computed from ground-truth simulations and model-generated trajectories.

For the stochastic Kuramoto--Sivashinsky (KS) equation, we use the joint probability density of the first and second spatial derivatives, $u_x$ and $u_{xx}$. Formally, if $\mu$ denotes the invariant measure over KS states $u$, the derivative-field joint density is:
\begin{equation}
    \rho_{\mathrm{KS}}(a,b)
    = \mathbb{E}_{u \sim \mu}
    \left[
    \frac{1}{L}\int_0^L
    \delta(a-u_x(x))\delta(b-u_{xx}(x))\,dx
    \right].
\end{equation}
This statistic summarizes the local phase-space geometry of KS rollouts. It is sensitive to changes in noise level and to failures where a surrogate converges to the wrong attractor, collapses variability, or distorts the relationship between slope and curvature. We compare the empirical joint densities from ground truth and surrogate rollouts using Wasserstein distance.

For the stochastic Kolmogorov flow, we use the time-averaged enstrophy spectrum. Let $\omega$ be the vorticity field and let $\widehat{\omega}(q)$ denote its Fourier coefficient at wavevector $q$. The shell-averaged enstrophy spectrum is:
\begin{equation}
    E_{\Omega}(k)
    = \mathbb{E}_{\omega \sim \mu}
    \left[
    \sum_{q:\, k \leq \|q\| < k+1}
    |\widehat{\omega}(q)|^2
    \right].
\end{equation}
This invariant measure statistic evaluates how vorticity intensity is distributed across spatial frequencies. It is especially important in low-viscosity regimes, where small-scale structures and high-frequency modes play a larger role in the dynamics. A surrogate that damps high frequencies,
injects spurious high-frequency artifacts, or misallocates enstrophy across scales will fail this metric even if individual snapshots appear visually plausible.

\subsection{Trustworthiness}

Forecast trustworthiness is especially important in machine learning for science because model outputs often inform downstream simulation, hypothesis generation, experimental design, and risk assessment. In these settings, an inaccurate uncertainty estimate can be as problematic as an inaccurate mean prediction. Overconfident surrogates may hide failure modes, while underconfident surrogates may be too diffuse to support useful decisions. We therefore evaluate probabilistic forecast quality using a scoring rule that rewards both calibration and sharpness.

We use the continuous ranked probability score (CRPS). For a predictive cumulative distribution $F$, samples $X,X' \sim F$, and observation $y$ and :
\begin{equation}
    \mathrm{CRPS}(F,y)
    =
    \int_{-\infty}^{\infty}
    \left(F(z) - \mathbf{1}\{y \leq z\}\right)^2 dz 
    =
    \mathbb{E}|X-y|
    -
    \frac{1}{2}\mathbb{E}|X-X'|.
\end{equation}

Lower CRPS indicates better probabilistic forecasts. The first term penalizes predictions that place mass far from the observation, while the second term rewards sharpness by penalizing unnecessary spread. This makes CRPS well suited for SPDE surrogate evaluation: it distinguishes models that are merely stochastic from models whose predictive distributions are statistically aligned with observed rollout errors. In our experiments, we estimate CRPS from sampled forecast trajectories and average over space, time, and test trajectories.

\subsection{Efficient Generative Scaling}
\label{sec:efficient-scaling}

We evaluate generative scaling primarily through wall-clock inference time, since probabilistic surrogates are useful in practice only if they can generate many rollout samples at reasonable cost. This is particularly important for stochastic interpolants, which provide high-quality conditional sampling but require numerical integration during inference. At full spatial resolution, each forecast step solves an interpolant SDE, making inference slower than a single deterministic neural-network forward pass. Stochastic interpolants \cite{albergo2025stochastic} define a continuous stochastic bridge between a current state $x(t)$ and a future state $x(t+\Delta t)$. We use interpolants of the form:
\begin{equation}
    I_s = \alpha_s \bm{x}(t) + \beta_s \bm{x}(t+\Delta t) + \sigma_s W_s,
    \qquad s \in [0,1],
\end{equation}
where $(\bm{x}(t), \bm{x}(t+\Delta t))$ is sampled from the empirical one-step transition distribution and $W_s$ is an independent Wiener process. The boundary conditions ensure that the bridge starts at the current state and ends at the next state. In our implementation: 
    $\alpha_s = 1-s, \quad
    \beta_s = s^2, \quad
    \sigma_s = 1-s, \quad
    W_s \overset{d}{=} \sqrt{s}z, \quad z \sim \mathcal{N}(0,I). $
At inference time, samples are generated by numerically integrating the learned interpolant dynamics. This direct stochastic bridge is well matched to autoregressive forecasting, but its cost grows with state dimension and the number of integration steps.

To reduce this cost, we forecast in a learned latent space. Recent work on latent generative models for physics \cite{rozet2025lost} has shown that autoencoder compression can substantially reduce sampling time. First, an encoder maps the full state $x$ to a latent representation $z = E(x)$ and a decoder reconstructs $\tilde{x} = D(z)$. We train an implicit rank-minimizing autoencoder \cite{jing2020implicit} with weight decay \cite{zeng2024autoencoders}. The stacked linear layers and weight decay implicitly promote a low-rank subspace with sharp singular value spectra cutoffs. This avoids sweeps over latent dimension with empirical metrics.

After training the autoencoder, we compute an SVD of the latent codes and retain the leading singular directions. The data are projected onto this discovered low-rank subspace, and the stochastic interpolant is trained on the projected latent states rather than the full-resolution fields. During generation, the interpolant samples in the reduced coordinates and the decoder maps the generated latent state back to the physical grid (for details see \ref{alg:dim-discovery}).

\section{Experiments and Data}

\subsection{Stochastic Kuramoto-Sivashinsky}
We begin by modeling the Kuramoto-Sivashinsky equation (KSE) (note change to spatial variable $x$ and state variable $u(x,t)$):
\begin{equation}
    \partial_t u  = -\partial_x^2 u - \partial_x^4 u - u \partial_x u + \eta(x,t)
\end{equation}
With periodic boundary conditions $u(0,t) =u(L,t)$. For appropriate choice of $L$ the deterministic KS exhibits a sustained chaotic attractor. The noise process is normally distributed $\eta \sim N(0,\sigma^2)$. We generate trajectories on domain sizes $L = \{22, 66\}$ and initial condition $u(0) \sim U(-0.4, 0.4)$ and use noise scales $\sigma = \{0.05, 0.5, 1.0, 2.0\}$. The stochastic KS has been widely studied as a model for physical systems with internal and external stochasticity \cite{gotoda2017chaotic}.

\subsection{Stochastic Kolmogorov}
The Kolmogorov 2D flow equation is a special case of the incompressible Navier-Stokes equations (NSE) with directional sinusoidal forcing:
\begin{equation}
    \partial_t \bm{u} + \nabla \cdot (\bm{u} \otimes \bm{u}) = \nu \nabla^2 \bm{u} - \frac{1}{\rho} \nabla p + \eta(t) \sin(ky)
\end{equation}
We add stochasticity to the system by drawing the amplitude of forcing from a Wiener process $\eta(t)$. We simulate 2D trajectories with jax-cfd \cite{dresdnerlearning} using periodic boundary conditions, a domain size of $[0, 2\pi]^2$, with initial conditions drawn from a random normal distribution adjusted to a maximum velocity 7.0, peak wavenumber 4, and zero divergence. We simulate 1000 trajectories and generate 3 datasets with viscosities $\nu = \{\num{1e-3}, \num{5e-4}, \num{1e-4}\}$. We use vorticity $\omega$ as the state variable for surrogate models, as is common practice in NSE surrogates.

For full SPDE details see Appendix \ref{app:spdes}.

\subsection{Experimental Setup}

We design our forecasting experiments for fair comparison between models and without specialized techniques. We assume improvements in temporal rollout schedule, physical regularization, and hard constraints would apply comparably to all models to facilitate evaluation of the underlying surrogate algorithm. Each model is trained for the same number of epochs, with the same data, and a similar number of parameters. During inference, all models begin with the initial condition and autoregressively rollout for the full trajectory length. See Appendix \ref{app:model-config} for full details. We compare the models:

\textbf{ResNets (ResNet)} are a popular benchmark. They are expressive, but require high memory due to full resolution layers with skip connections requiring many activations to be stored.

\textbf{UNets (UNet)} are another popular benchmark. We use sinusoidal positional embeddings and 4 spatial attention layers. \textbf{UNet M} is the UNet model with Monte Carlo dropout \cite{gal2016dropout} and \textbf{UNet P} is the UNet model trained with a heteroscedastic Gaussian negative log likelihood \cite{nix1994estimating}.

\textbf{PDE-Transformer (PDE-T)} \cite{holzschuh2025pde} is a recent transformer based model specifically designed for forecasting PDEs. This variant uses MSE training.

\textbf{PDE-Transformer Flow (PDE-TF)} learns an ODE based conditional flow map between time steps. In contrast to the multi-step rollout in \cite{holzschuh2025pde}, we predict only one step ahead given the current state for fair comparison.

\textbf{Stochastic Interpolants (SI)} described in Section \ref{sec:efficient-scaling} use a ResNet for 1D data and a UNet for 2D data.

\textbf{Latent Stochastic Interpolants (Latent SI)} uses an implicit rank-minimizing autoencoder (IRMAE) with weight decay and SVD projection to automatically discover the low-rank latent space of the data. After training the IRMAE, we train Latent SI on to forecast in this low-rank space. See Algorithm \ref{alg:dim-discovery} for full details.

\section{Results and Discussion}

\begin{table*}[ht]
 \caption{Wasserstein distance of predictions vs ground truth for invariant measures (full dimension).}
 \vskip 0.05in
 \label{tab:main-norm}
 \centering
 \footnotesize
     \begin{sc}
       \begin{tabular}{lccccccc}
           \toprule
           Dataset & SI & ResNet & UNet & PDE-T & PDE-TF & UNet M & UNet P \\ \midrule
           KS $L=22 \ \sigma=0.05$ & 0.163 & 0.156 & 0.203 & \textbf{0.161} & 1.451 & 0.715 & 0.821 \\
           KS $L=22 \ \sigma=0.5$ & \textbf{0.188} & 0.296 & 0.221 & 0.253 & 0.212 & 0.816 & 0.974 \\
           KS $L=22 \ \sigma=1.0$ & \textbf{0.311} & 0.438 & 0.324 & 0.387 & 0.340 & 1.370 & 1.520 \\
           KS $L=22 \ \sigma=2.0$ & \textbf{0.514} & 1.366 & 1.229 & 6.282 & 0.573 & 2.118 & 2.801 \\
           KS $L=66 \ \sigma=0.05$ & 0.125 & \textbf{0.123} & 0.127 & 0.249 & 0.181 & --- & 0.896 \\
           KS $L=66 \ \sigma=0.5$ & \textbf{0.130} & 0.161 & 0.172 & 0.267 & 0.148 & 0.806 & 0.749 \\
           KS $L=66 \ \sigma=1.0$ & \textbf{0.150} & 0.443 & 0.435 & 0.457 & 0.158 & 1.025 & 1.126 \\
           KS $L=66 \ \sigma=2.0$ & \textbf{0.226} & 1.189 & 0.975 & 0.576 & 0.282 & 1.316 & 1.882 \\
           Kolmogorov $\nu = \num{1e-3}$ & \textbf{0.00107} & 0.122 & 0.00178 & 0.0126 & 0.00215 & 0.00545 & 0.14979 \\
           Kolmogorov $\nu = \num{5e-4}$ & 0.00158 & 0.165 & 0.00168 & 0.00141 & \textbf{0.00062} & 0.00319 & 0.21401 \\
           Kolmogorov $\nu = \num{1e-4}$ & \textbf{0.00068} & 0.163 & 0.00117 & 0.0127 & 0.00076 & 0.08090 & 0.16890 \\
           \bottomrule
       \end{tabular}
     \end{sc}
\end{table*}

We evaluate neural surrogates using the criteria introduced in Section 3. First, a surrogate should reproduce the long-time invariant measures of the system. Second, forecast distributions should be well calibrated. Third, generation should be efficient to mitigate cost relative to deterministic models. Across these criteria, SIs provide the most consistent performance, particularly in high-noise regimes.

\begin{figure*}[ht]
  %\vskip 0.2in
  \centering
    \includegraphics[width=\textwidth]{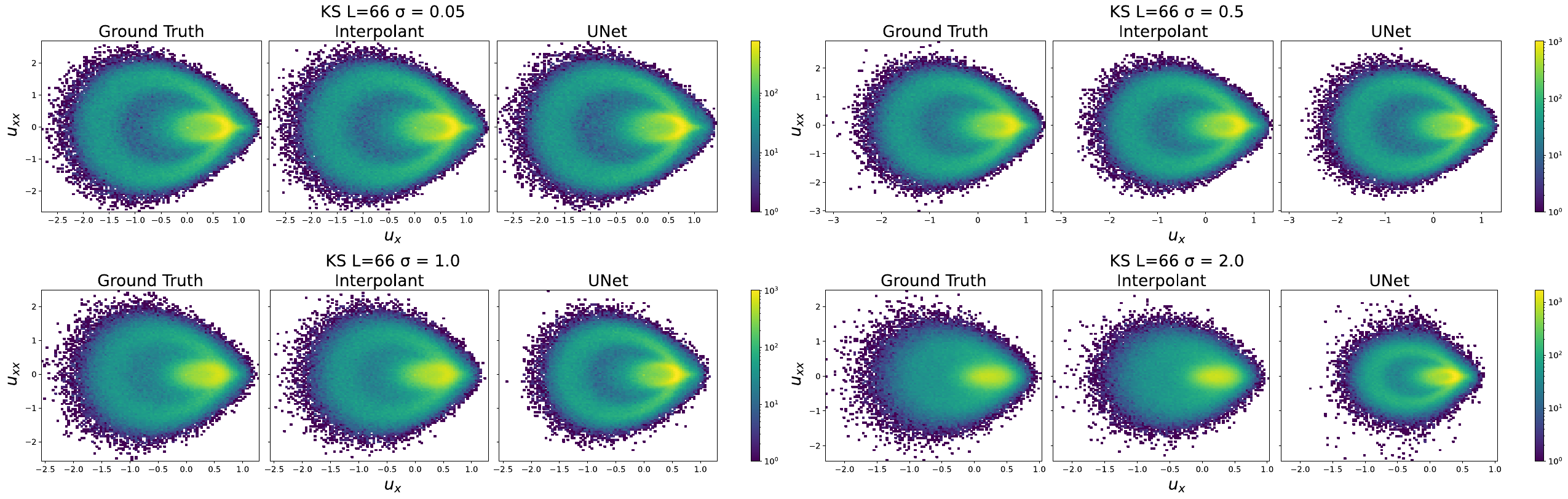}
    \caption{
      Joint PDF of KSE $u_x$ vs $u_{xx}$ test data, domain size $L=66$. The interpolant captures smoothing of the peak features with increasing $\sigma$ which the deterministic model does not.
    }
    \label{fig:ks-joint-1}
\end{figure*}

\begin{figure}[ht]
\centering
\begin{subfigure}{.55\columnwidth}
  \centering
  \includegraphics[width=\linewidth]{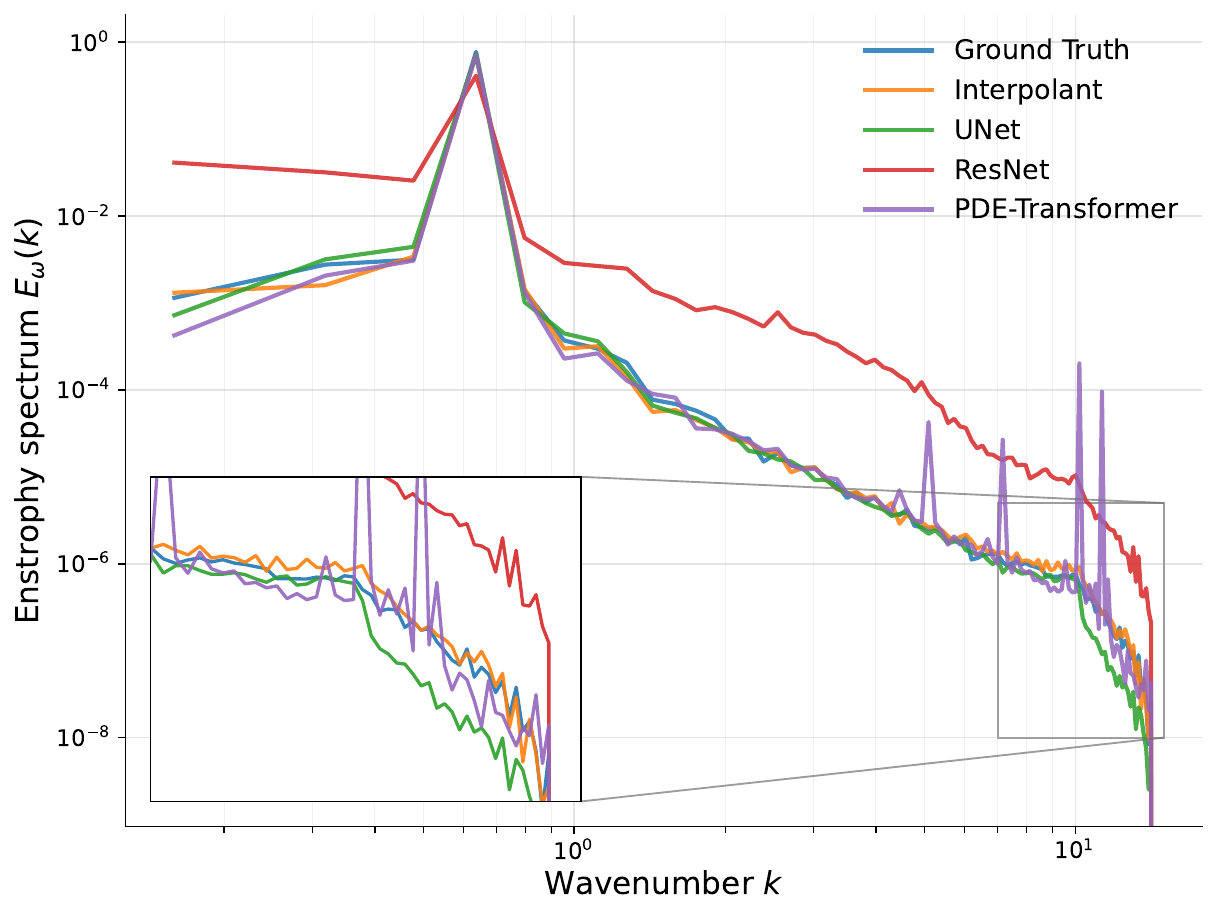}
  \caption{}
  \label{fig:kol-ens-preds-a}
\end{subfigure}%
\begin{subfigure}{.45\columnwidth}
  \centering
  \includegraphics[width=\linewidth]{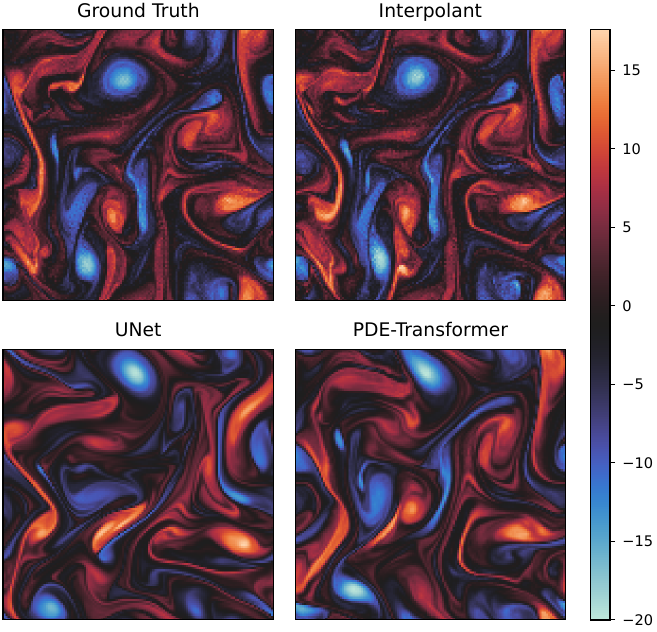}
  \caption{}
  \label{fig:kol-ens-preds-b}
\end{subfigure}
\caption{(a) Time averaged enstrophy spectrum of 2D Kolmogorov low viscosity $\nu = 1e-4$ test trajectories vs model prediction enstrophy spectra. (b) 2D Kolmogorov sample predictions at $t=35$. Note that the Interpolant matches the ground truth spectrum best, especially at high frequencies shown in the popout box.}
\label{fig:kol-ens-preds}
\end{figure}

\subsection{Can Neural Surrogates Accurately Capture Invariant Measures?}

Table \ref{tab:main-norm} and Figures \ref{fig:ks-joint-1} and \ref{fig:kol-ens-preds} compare the invariant measure statistics of autoregressive surrogate rollouts against ground-truth SPDE simulations. The first clear finding is that pointwise-trained neural surrogates fail for all architectures. In contrast, SIs are consistently among the most accurate models and are especially robust in high-noise and low-viscosity regimes.

In the KSE low noise regime (Figure \ref{fig:ks-joint-1}), deterministic UNet and SI models both accurately preserve the invariant measure because the dynamics remain close to the deterministic chaotic attractor. As the noise scale increases, the deterministic UNet performs poorly. This occurs because the model lacks the stochastic component required to capture switching between the deterministic chaotic attractor and the emergent noise dominated regime. For an individual time series visualization, see Figure \ref{fig:ks-66-sample-1}. The SI tracks captures the emergent noisy regime and dynamic swtiching more faithfully, as seen qualitatively in Figures \ref{fig:ks-joint-1},\ref{fig:ks-66-sample-1}  and quantitatively by the Wasserstein distances in Table \ref{tab:main-norm}. See App. \ref{app:further-results} for all model results. All deterministic models exhibit similar behavior, while both SI and PDE-TF perform well.

The Kolmogorov experiments test preservation of the vorticity distribution across spatial scales. In the low-viscosity case  (Figure \ref{fig:kol-ens-preds-a}), the SI most closely matches the ground-truth enstrophy spectrum, particularly at high frequencies. The deterministic baselines show distinct failure modes. The ResNet is incorrect across multiple frequencies, the UNet damps small-scale activity, and the PDE-Transformer exhibits spike-like spectral artifacts. As shown in Figure \ref{fig:kol-ens-preds-b}, these spectral errors correspond to visible differences: after autoregressive rollout, the SI better preserves coherent vortical structures while the baselines either oversmooth or introduce spurious spatial structure.

Comparing the SI and the flow-based PDE-Transformer shows the choice of probabilistic model is important. Flow matching is competitive, but less consistent than SI. A likely reason is that the SI learns a direct stochastic bridge between the current and next states, whereas the conditional flow transports noise through an ODE. The SI's SDE-based inference injects stochasticity throughout generation, which appears to improve long-time stability and preserve invariant measure structure during rollout. Both generative models clearly outperform Monte Carlo Dropout and Gaussian NLL trained models by the invariant measure metric. We speculate this is due to accumulation of error from improper variance structure in the latter models.

\begin{table}[ht]
  \caption{Mean CRPS of model predictions over full test trajectories.}
  \label{tab:crps}
  \centering
  \small
      \begin{sc}
        \begin{tabular}{lccc}
            \toprule
            Dataset & SI & UNet M & UNet P \\ \midrule
            KS $L=66 \ \sigma=0.05$ & 1.15 & --- & 1.23 \\
            KS $L=66 \ \sigma=0.5$ & 1.22 & 1.35 & 1.26 \\
            KS $L=66 \ \sigma=1.0$ & 1.47 & 4.02 & 1.49 \\
            KS $L=66 \ \sigma=2.0$ & 2.17 & 5.14 & 2.94 \\
            Kolmogorov $\nu = \num{1e-3}$ & 1.59 & 1.84 & 2.11 \\
            Kolmogorov $\nu = \num{5e-4}$ & 1.69 & 2.01 & 2.02  \\
            Kolmogorov $\nu = \num{1e-4}$ & 1.80 & 2.65 & 2.19 \\
            \bottomrule
        \end{tabular}
      \end{sc}
\end{table}

\begin{figure}[ht]
\centering
\begin{subfigure}{.3\textwidth}
  \centering
  \includegraphics[width=\linewidth]{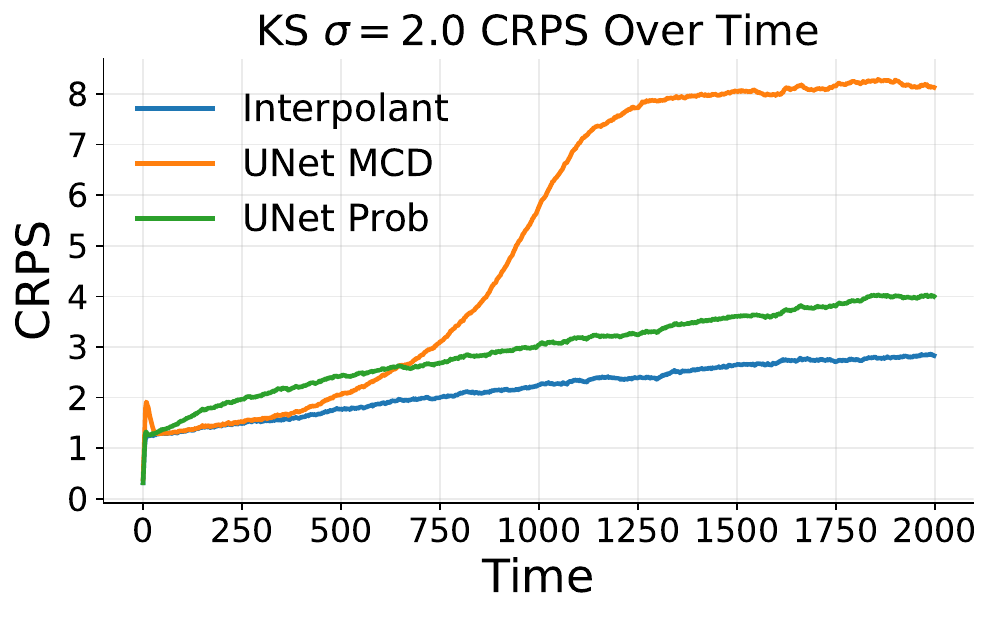}
  \caption{}
  \label{fig:ks-crps-a}
\end{subfigure}%
\begin{subfigure}{.3\textwidth}
  \centering
  \includegraphics[width=\linewidth]{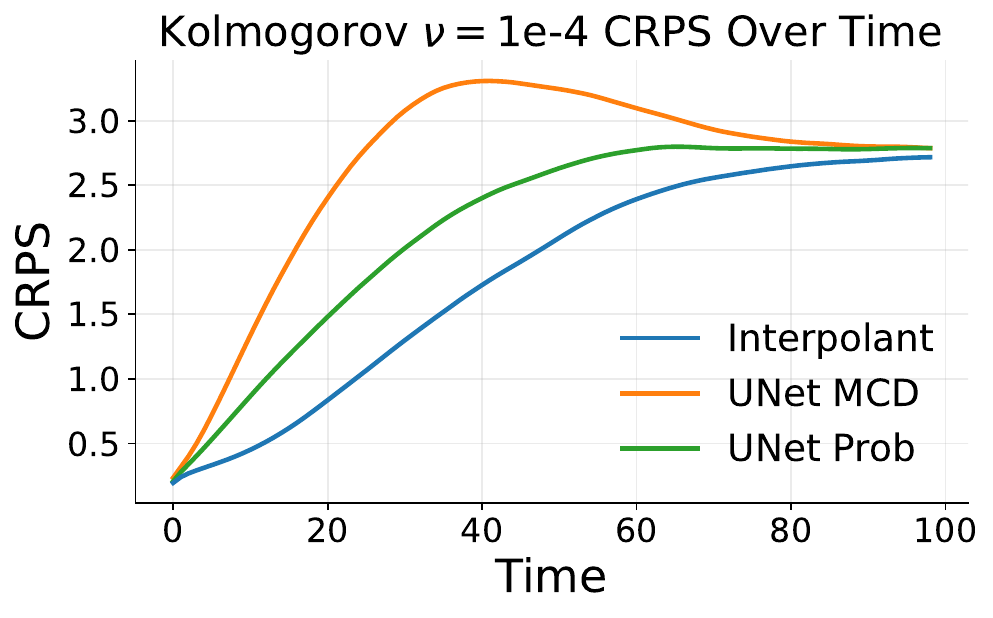}
  \caption{}
  \label{fig:ks-crps-b}
\end{subfigure}
\caption{KS and Kolmogorov Rollout CRPS over time.}
\label{fig:ks-crps}
\end{figure}

\begin{figure}[ht]
    \centering
    \includegraphics[width=0.8\columnwidth]{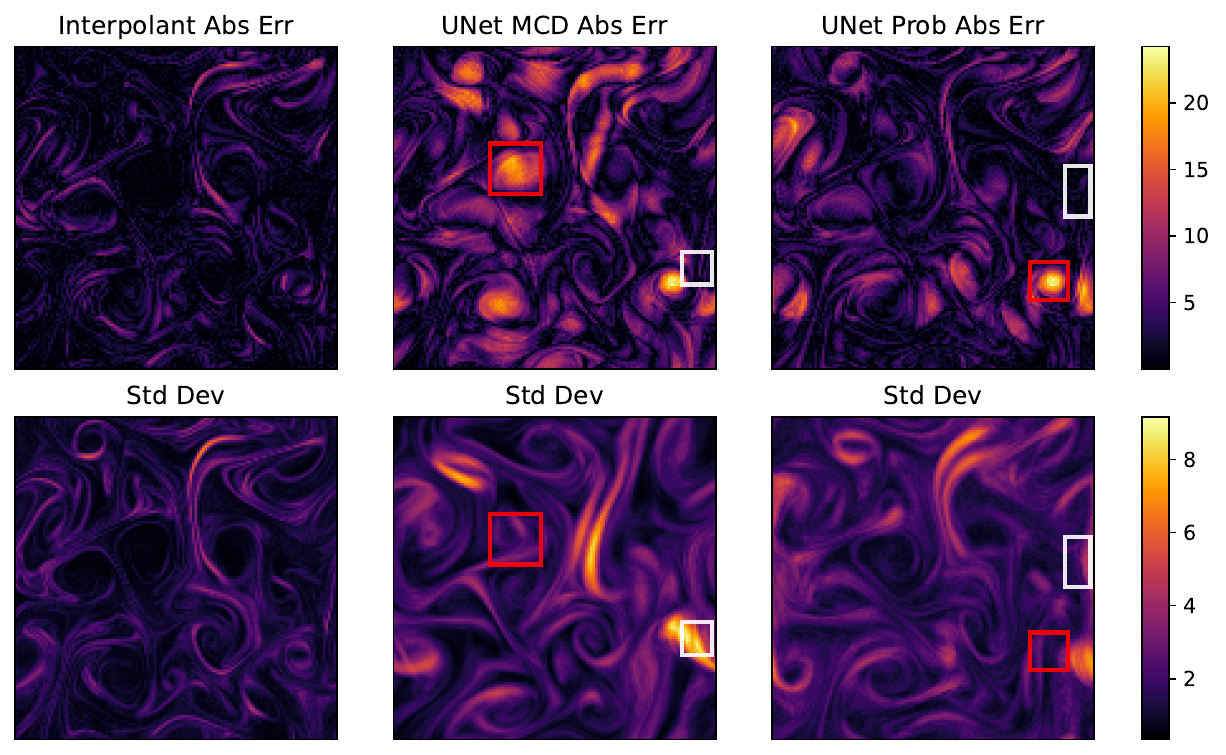}
    \caption{Absolute error and standard deviation (uncertainty) for a Kolmogorov trajectory at timestep 30. Note that the Interpolant uncertainty matches the absolute error closely while UNet MCD and UNet Prob both suffer from over-confident regions (red boxes) and under-confident regions (white boxes).}
    \label{fig:kol-unc-regions}
\end{figure}

\subsection{How Trustworthy are Surrogate Predictions?}

Table \ref{tab:crps} and Figures \ref{fig:ks-crps} and \ref{fig:kol-unc-regions} evaluate whether probabilistic surrogates produce forecast distributions that are calibrated and sharp. The SI achieves the lowest mean CRPS in every reported setting, indicating that its predictive distributions better match the empirical rollout errors. Figure 3a shows the accumulation of this effect over time. The SI's CRPS grows gradually, while the dropout and NLL baselines accumulate larger error. However, NLL forecasts appear acceptable considering the modest increase in CRPS relative to ease of training. To emphasize the difference from generative models, we recall the perform invariant measure performance and consider a more detailed uncertainty metric.

The Kolmogorov results show a more localized but still important calibration failure. In Figure \ref{fig:ks-crps-b}, the CRPS curves for the UNet baselines rise sharply at intermediate times before eventually approaching the SI. This indicates that their forecast distributions are poorly matched to the actual uncertainty during the most dynamically sensitive part of the rollout. Figure \ref{fig:kol-unc-regions} makes this failure spatially explicit. A trustworthy model should assign high uncertainty where its absolute error is high and lower uncertainty elsewhere. The interpolant's standard-deviation field follows this pattern more closely, while the UNet baselines contain both overconfident regions, where the error is large but the uncertainty is small, and underconfident regions, where the uncertainty is large without corresponding error.

These results show that approximate stochastic output is not enough for trustworthy SPDE forecasting. Monte Carlo dropout and heteroscedastic Gaussian likelihoods produce uncertainty estimates, but those estimates are not consistently aligned with actual rollout error. SIs provide better calibrated forecast distributions because they directly model the conditional transition distribution rather than adding uncertainty to a pointwise predictor.

\subsection{Can Probabilistic Surrogates Be Scaled for Efficient Generation?}

The previous sections show that SIs improve invariant measure fidelity and forecast trustworthiness, but full-dimensional sampling is computationally expensive. Table \ref{tab:low-dim-norm} evaluates invariant measure reproduction for SIs in the IRMAE \cite{jing2020implicit} latent space. For KS, the latent SI remains close to the full-dimensional SI across noise levels. There is noticeable degradation due to latent compression in the Kolmogorov case, as expected due to the stronger scale coupling present in the NSE relative to the KS. However, the predictions remain stable and near the true attractor. We include latent ResNet as a baseline to verify conclusions from the full dimension results. While performance remains poor relative to the SI, invariant measure reproduction actually improves for high noise KSE and Kolmogorov flow.

For a single Kolmogorov state, full-dimensional SI inference takes 1.709 seconds, compared to 0.141 in the latent space, or a speedup of $\approx 12 \times$. The speedup arises because SDE integration is performed in a compressed coordinate system rather than on the full spatial grid. These results demonstrate a accuracy--cost tradeoff. Full-dimensional stochastic interpolants provide statistical fidelity, while latent stochastic interpolants accelerate inference with some loss in accuracy. This makes probabilistic SPDE forecasting more practical for long rollouts and repeated sampling, where generation speed is a central bottleneck.

\begin{table}[ht]
  \caption{Wasserstein Distance of Predictions vs Ground Truth for Invariant Measures with reduced dimension models.}
  \label{tab:low-dim-norm}
  \centering
      \begin{sc}
        \begin{tabular}{lccccc}
            \toprule
            Dataset & Latent SI & SI & Latent ResNet & ResNet \\ \midrule
            KS $L=66 \ \sigma=0.05$ & 0.125 & 0.125 & 0.578 & 0.123 \\
            KS $L=66 \ \sigma=0.5$ & 0.121 & 0.130 & 0.501 & 0.161 \\
            KS $L=66 \ \sigma=1.0$ & 0.176 & 0.150 & 0.586 & 0.443 \\
            KS $L=66 \ \sigma=2.0$ & 0.275 & 0.226 & 0.889 & 1.189  \\
            Kolmogorov $\nu = \num{1e-3}$ &  0.00692 & 0.00107 & 0.0214 & 0.122 \\
            Kolmogorov $\nu = \num{5e-4}$ & 0.00529 & 0.00158 & 0.0199 & 0.165 \\
            Kolmogorov $\nu = \num{1e-4}$ & 0.0103 & 0.00068 & 0.0315 & 0.163 \\
            \bottomrule
        \end{tabular}
      \end{sc}
\end{table}

\subsection{Limitations and Future Work}
\label{sec:limitations}

A significant limitation is the stationary assumption. For transient dynamics, invariant measures may not exist or meaningfully characterize the dynamics. Alternative measures related to transient chaos \cite{grebogi1983crises}, chaotic transients \cite{kantz1985repellers}, and almost invariant sets \cite{pianigiani1979expanding} may be applicable, but these have received significantly less attention and development. Transient systems generally will not admit global reduced order representations, and even stationary systems are not guaranteed to converge to an attractor. This will highlight areas where further research is required for accurate, efficient reduced order generation. Even for systems speculated to live on invariant attractors, length scale coupling may be strong enough to limit accurate dimension reduction. Other effects neglected in the current data sets, but can be addressed by the framework include: multiplicative and state-dependent noise, noisy or partial observations, and more generic forms of uncertainty.

\section{Conclusion}

Reliable surrogate forecasting for stochastic PDEs requires evaluation beyond short-horizon pointwise error. We introduced TRIE, a framework that evaluates whether SPDE surrogates reproduce invariant measures, provide trustworthy predictive uncertainty, and scale to efficient probabilistic generation. Across stochastic Kuramoto-Sivashinsky and stochastic Kolmogorov flow, TRIE reveals that pointwise-trained surrogates can produce plausible rollouts while failing to match long-time statistical structure, and that approximate uncertainty methods can be miscalibrated or overconfident. In contrast, distributional generative surrogates, particularly stochastic interpolants, provide the most consistent performance across the TRIE criteria, achieving strong invariant measure fidelity and the lowest CRPS across all reported probabilistic settings. Finally, latent stochastic interpolants retain much of this statistical fidelity while reducing Kolmogorov inference time by roughly $12\times$, demonstrating a practical accuracy-cost tradeoff. We release datasets and code to support reproducible evaluation and further development of trustworthy stochastic PDE forecasting methods.

\begin{ack}
This research used resources provided by the Los Alamos National Laboratory (LANL) Institutional Computing Program and the Darwin testbed at LANL, which is funded by the Computational Systems and Software Environments subprogram of LANL’s Advanced Simulation and Computing program (NNSA/DOE). This work was supported by the LANL Laboratory Directed Research and Development under the Center for Nonlinear Studies program 20250614CR-NLS. LANL is operated by Triad National Security, LLC, for the National Nuclear Security Administration of the U.S. Department of Energy (Contract No. 89233218CNA000001).
\end{ack}

\bibliographystyle{plainnat}
\bibliography{refs}

\appendix

\section{Technical appendices and supplementary material}

\subsection{SPDEs and Simulation}
\label{app:spdes}

\subsubsection{Stochastic Kuramoto-Sivashinsky}

We model the Kuramoto-Sivashinsky equation (KSE) (note change to spatial variable $x$ and state variable $u(x,t)$):
\begin{equation*}
    \partial_t u  = -\partial_x^2 u - \partial_x^4 u - u \partial_x u + \eta(x,t)
\end{equation*}
With periodic boundary conditions $u(0,t) =u(L,t)$. For appropriate choice of $L$ the deterministic KS exhibits a sustained chaotic attractor. The noise process is normally distributed $\eta \sim N(0,\sigma^2)$ with $\langle \eta(x,t) \rangle = 0$ and $\langle \eta(x,t) \eta(x',t') \rangle = 2\delta(x-x') \delta(t-t')$. We generate trajectories on domain sizes $L = \{22, 66\}$, simulation step size $dt = 0.05$, training time step size $\Delta t = 0.25$, and initial condition $u(0) \sim U(-0.4, 0.4)$. We remove the first 10000 steps of each trajectory corresponding to transience time $\tau = 500$. For KS $L = 22$, we discretize the state space into $N = 64$ grid points and use noise scales $\sigma = \{0.05, 0.5, 1.0, 2.0\}$, generating 1000 trajectories for each. For KS $L = 66$, we discretize the state space into $N = 128$ points and use the same noise scales. The stochastic KS has been widely studied as a model for physical systems with internal and external stochasticity. For $\sigma \lessapprox 0.5$, the deterministic terms dominate and the dynamics approximately follow the chaotic attractor \cite{gotoda2017chaotic}. For stronger noise, $\sigma=\{1.0,2.0\}$, the dynamics transition between the chaotic attractor and noise-dominated stochastic dynamics. We do not study the noise-dominated regime obtained by further increasing $\sigma$.

\subsubsection{Stochastic Kolmogorov}

The Kolmogorov 2D flow equation is a special case of the incompressible Navier-Stokes equations (NSE) with directional sinusoidal forcing:
\begin{equation*}
    \partial_t \bm{u} + \nabla \cdot (\bm{u} \otimes \bm{u}) = \nu \nabla^2 \bm{u} - \frac{1}{\rho} \nabla p + \eta(t) \sin(ky)
\end{equation*}
We add stochasticity to the system by drawing the amplitude of forcing from a Wiener process $\eta(t)$. We simulate 2D trajectories with jax-cfd \cite{dresdnerlearning} using periodic boundary conditions, a domain size of $[0, 2\pi]^2$, a grid size of $N = 256 \times 256$, with initial conditions drawn from a random normal distribution adjusted to a maximum velocity 7.0, peak wavenumber 4, and zero divergence. We simulate 1000 trajectories using a simulation timestep of $dt = 0.001$ and remove the first 100,000 steps corresponding to $\tau = 100$. We generate 3 datasets with viscosities $\nu = \{\num{1e-3}, \num{5e-4}, \num{1e-4}\}$ and a training timestep of $\Delta t = 0.1$. We use vorticity $\omega$ as the state variable for surrogate models, as common practice in NSE surrogates. This is the first implementation of a stochastic Kolmogorov flow solver we could find, although there are comparable studies of periodically forced 2D NSE with random forcing \cite{hairer2006ergodicity}.

\subsection{Model Configuration and Training}
\label{app:model-config}

\subsubsection{Stochastic 1D Kuramoto-Sivashinsky Training Details}

We train each model for 2000 epochs using the AdamW optimizer with a learning rate of 1e-3. We use 800 KS trajectories for training and 150 trajectories for our test set. Each trajectory has 2000 timesteps. For this dataset, the stochastic interpolant uses a ResNet architecture as its backbone.

\subsubsection{Stochastic 2D Kolmogorov Training Details}

We train each model for 100 epochs using the AdamW optimizer with a learning rate of 2e-4. Models are trained with 800 trajectories with 100 timesteps each. We hold out 150 trajectories as a test set. In this case, the stochastic interpolant uses a UNet architecture as its backbone.

\subsubsection{Hardware}

Our models were trained using 4 Nvidia A100-SXM4 40 GB gpus. Inference jobs were run on a single A100 gpu.

\subsection{Further Results}
\label{app:further-results}

\begin{figure}[ht]
  \centering
    \includegraphics[width=\columnwidth]{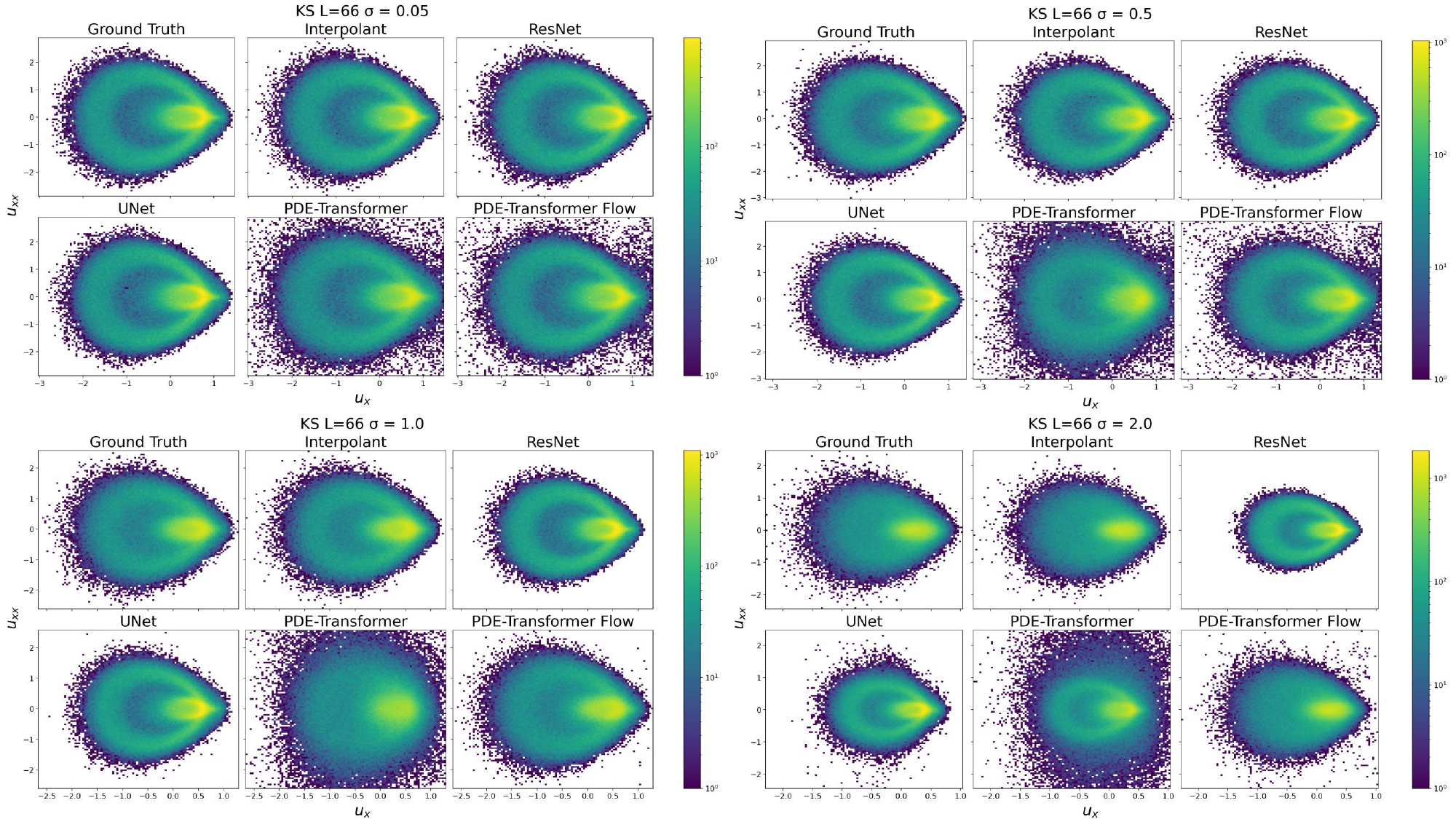}
    \caption{
      Joint PDF of Kuramoto-Sivashinsky for all models trained on domain size $L=66$.
    }
    \label{fig:ks-66-joint-all}
\end{figure}

\begin{figure}[ht]
  \centering
    \includegraphics[width=0.8\columnwidth]{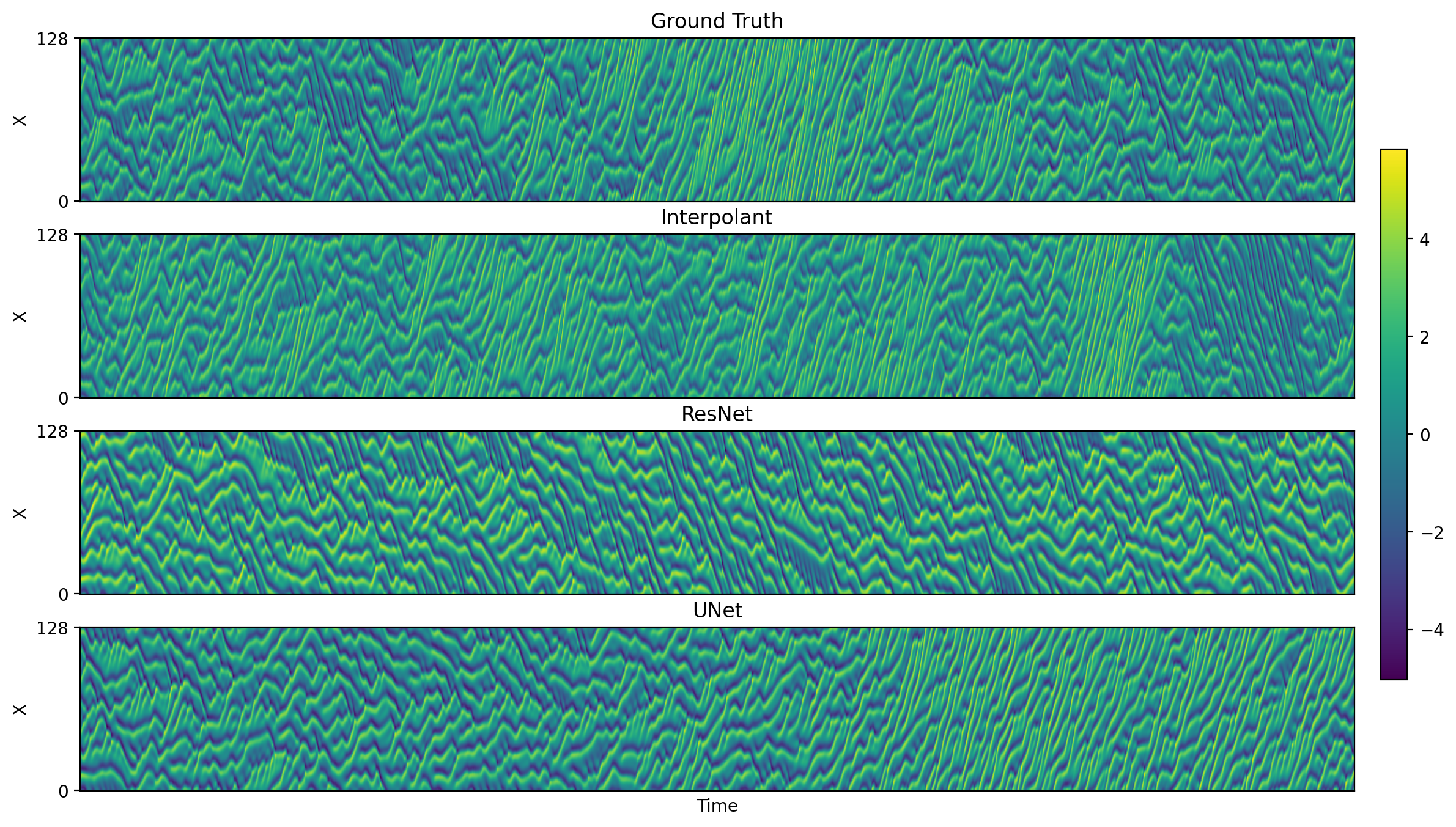}
    \caption{
      Spacetime plot of Kuramoto-Sivashinsky test trajectory with domain size $L=66$ and noise scale $\sigma=1.0$.
    }
    \label{fig:ks-66-sample-1}
\end{figure}

Figure \ref{fig:ks-66-joint-all} illustrates the joint PDF for all the main models in this paper. Note that the stochastic interpolant visually matches the ground truth best, especially in the high noise cases. Figure \ref{fig:ks-66-sample-1} illustrates KS model rollouts for a $\sigma=1.0$ noise trajectory.

\begin{figure}[ht]
\centering
\begin{subfigure}{.33\textwidth}
  \centering
  \includegraphics[width=\linewidth]{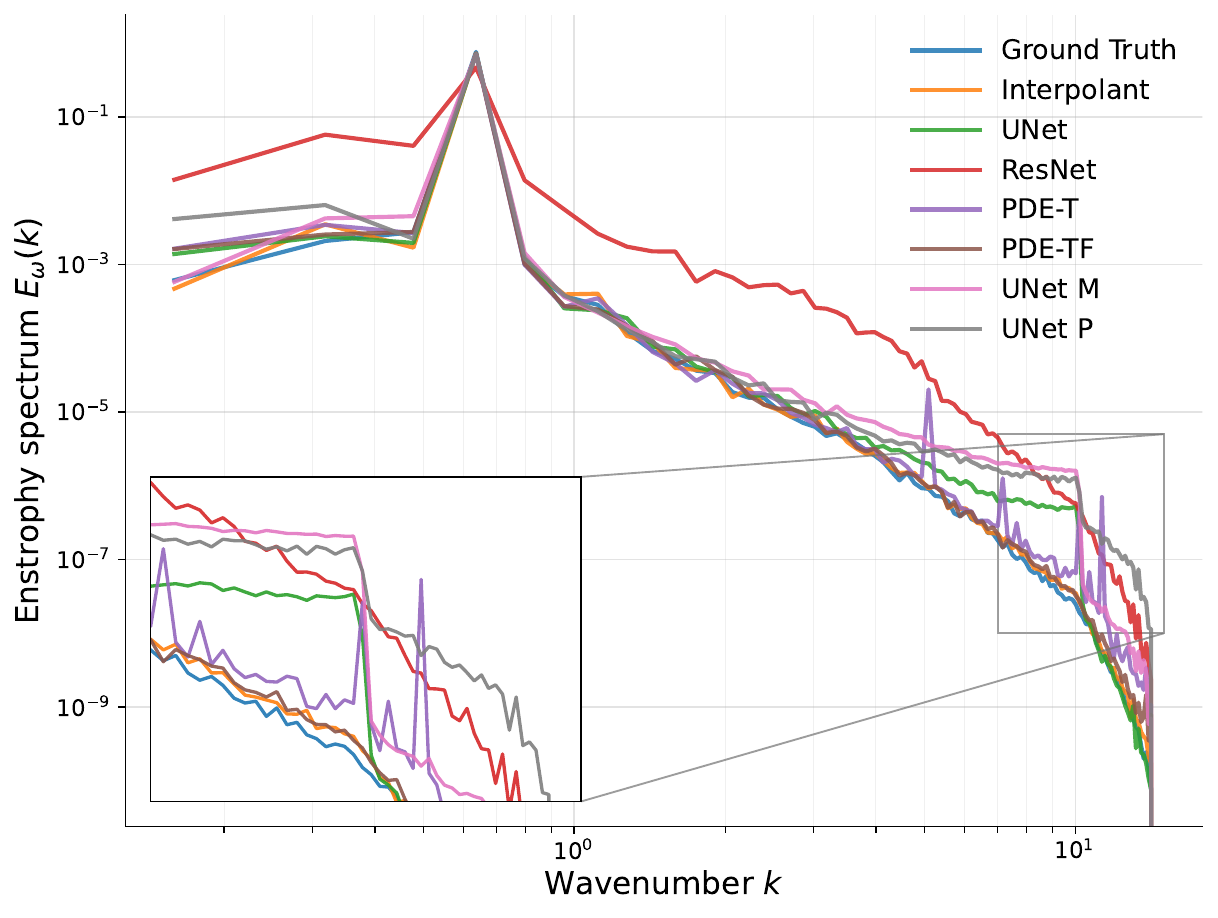}
  \caption{Viscosity 1e-3}
  \label{fig:kol-ens-all-a}
\end{subfigure}%
\begin{subfigure}{.33\textwidth}
  \centering
  \includegraphics[width=\linewidth]{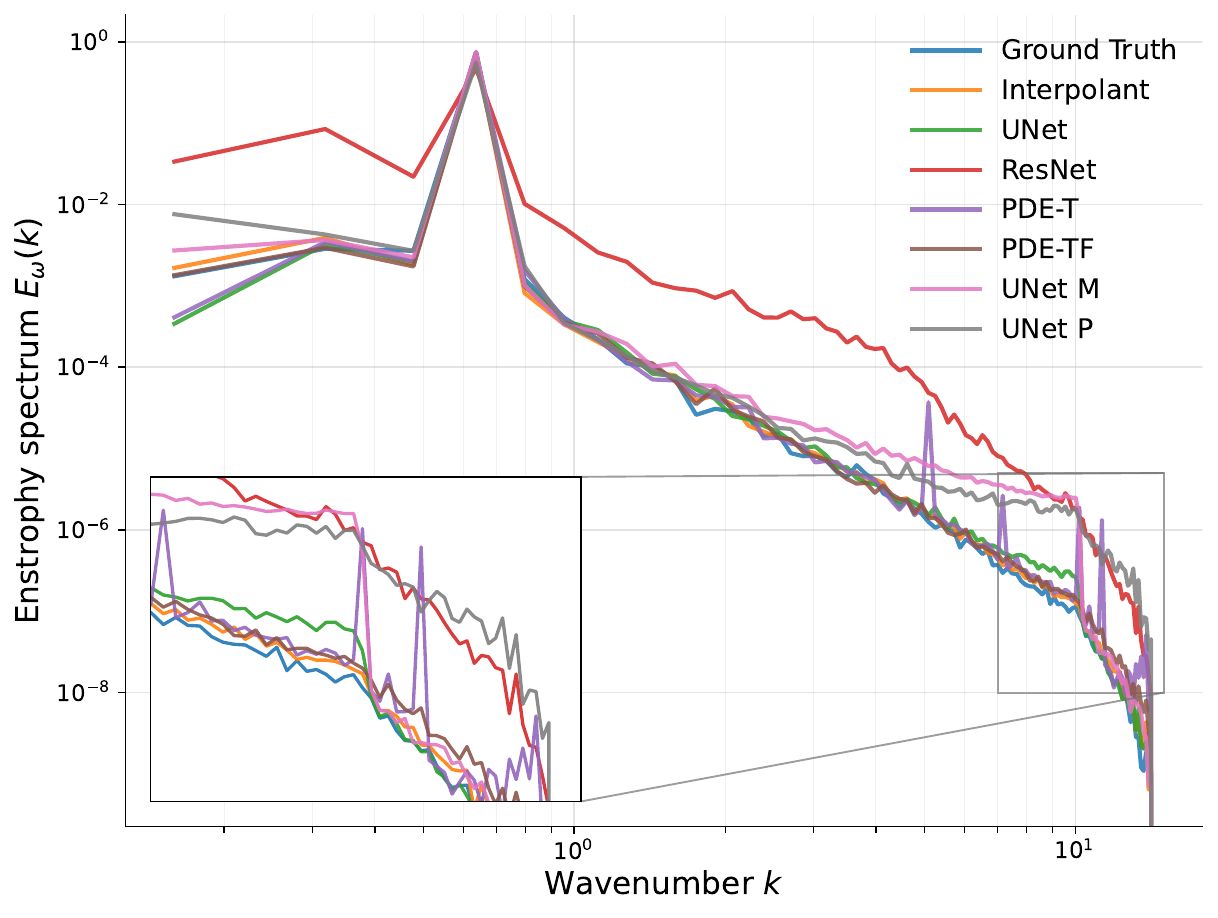}
  \caption{Viscosity 5e-4}
  \label{fig:kol-ens-all-b}
\end{subfigure}
\begin{subfigure}{.33\textwidth}
  \centering
  \includegraphics[width=\linewidth]{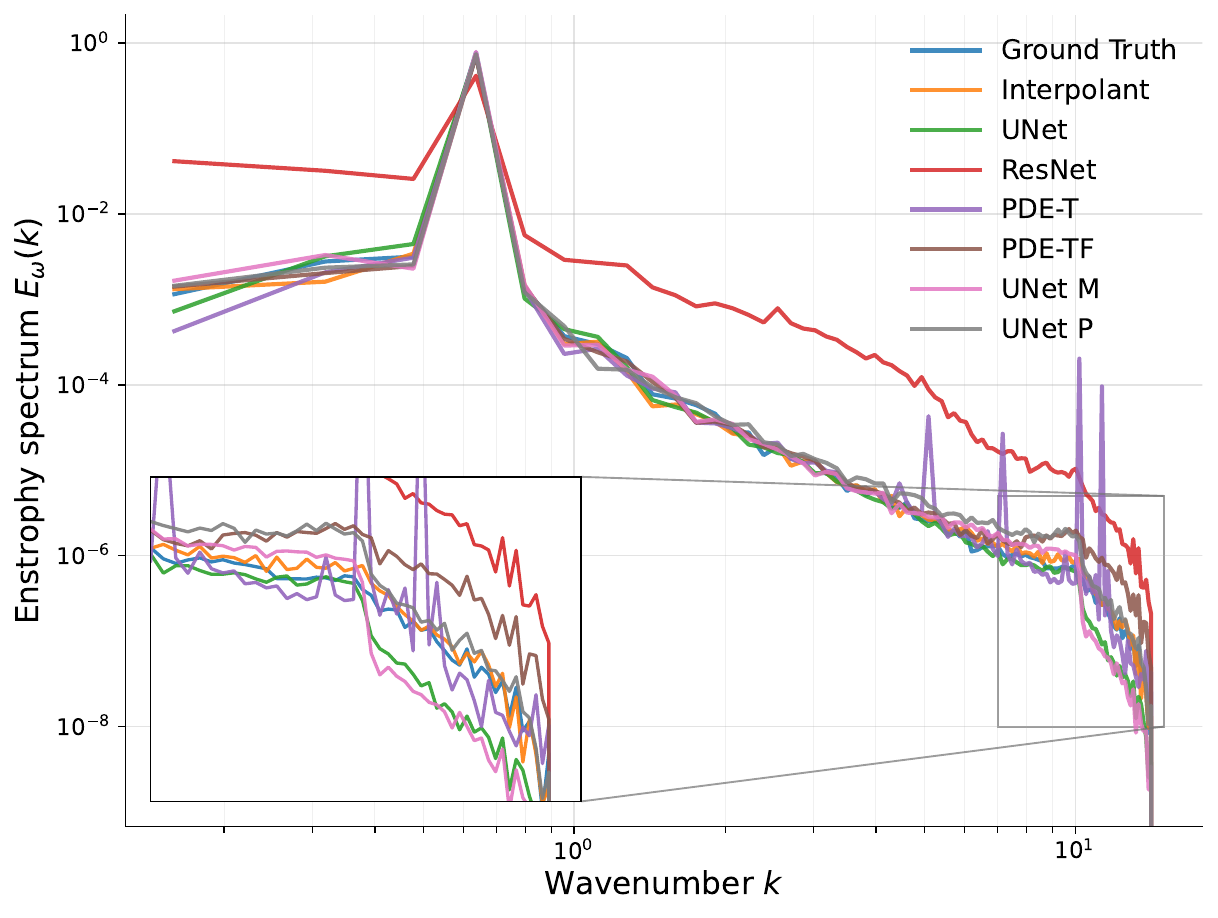}
  \caption{Viscosity 1e-4}
  \label{fig:kol-ens-all-c}
\end{subfigure}
\caption{Kolmogorov enstrophy spectra for all models.}
\label{fig:kol-ens-all}
\end{figure}

\begin{figure}[ht]
  \centering
    \includegraphics[width=0.89\columnwidth]{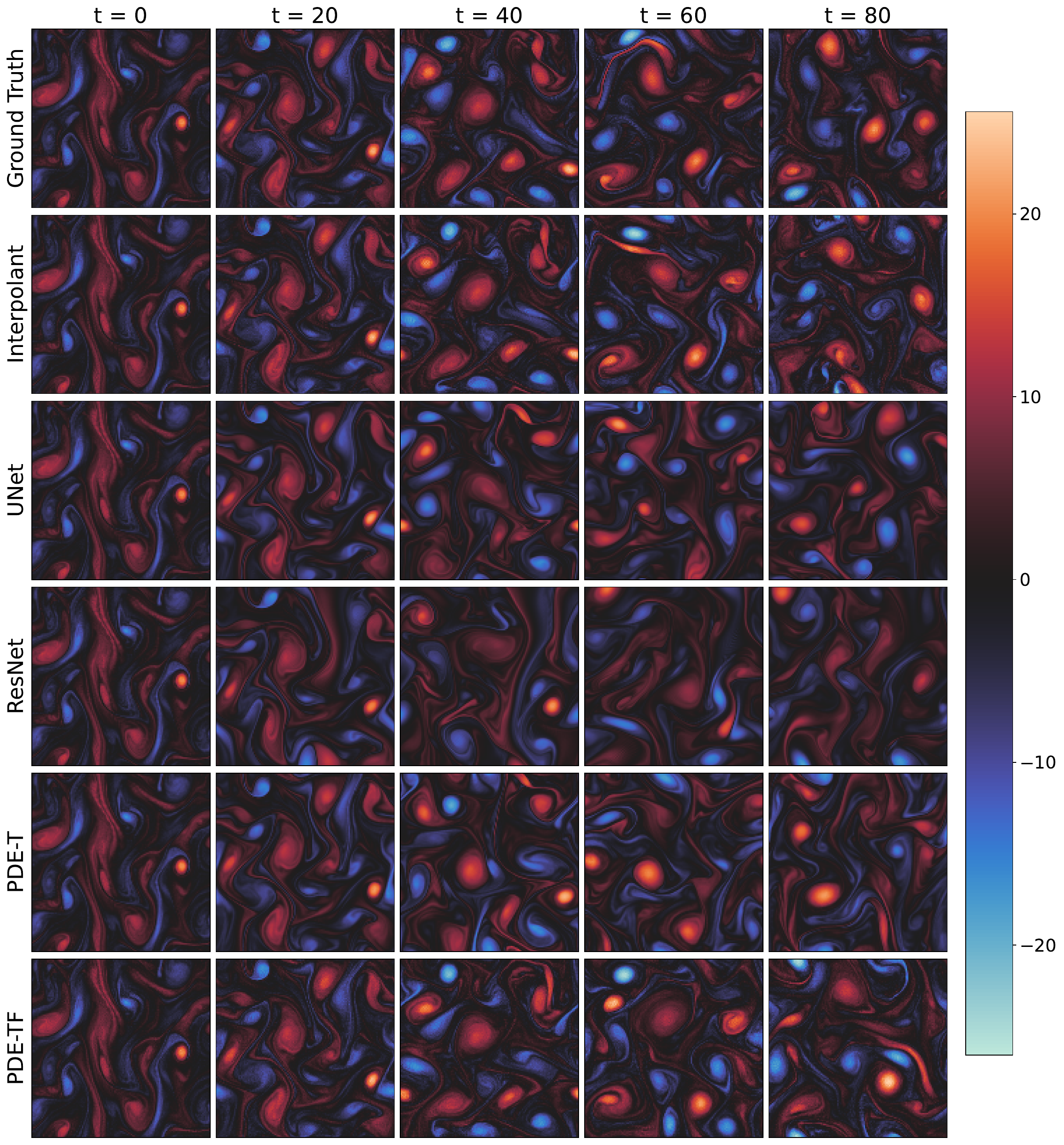}
    \caption{
      Model rollouts vs Ground Truth of Kolmogorov test trajectory with viscosity $\nu = 1e-4$.
    }
    \label{fig:kol-sample-1}
\end{figure}

Figure \ref{fig:kol-ens-all} shows the Kolmogorov enstrophy spectra for all 3 viscosities in this study with high frequencies highlighted. The interpolant and PDE-TF models match the ground truth spectra best. Figure \ref{fig:kol-sample-1} shows model rollouts for a Kolmogorov $\nu=1e-4$ trajectory vs ground truth.

\newpage

\subsection{Latent Dimension Discovery}
\label{app:dim-discovery}

\begin{figure}[ht]
\centering
\begin{subfigure}{.47\textwidth}
  \centering
  \includegraphics[width=\linewidth]{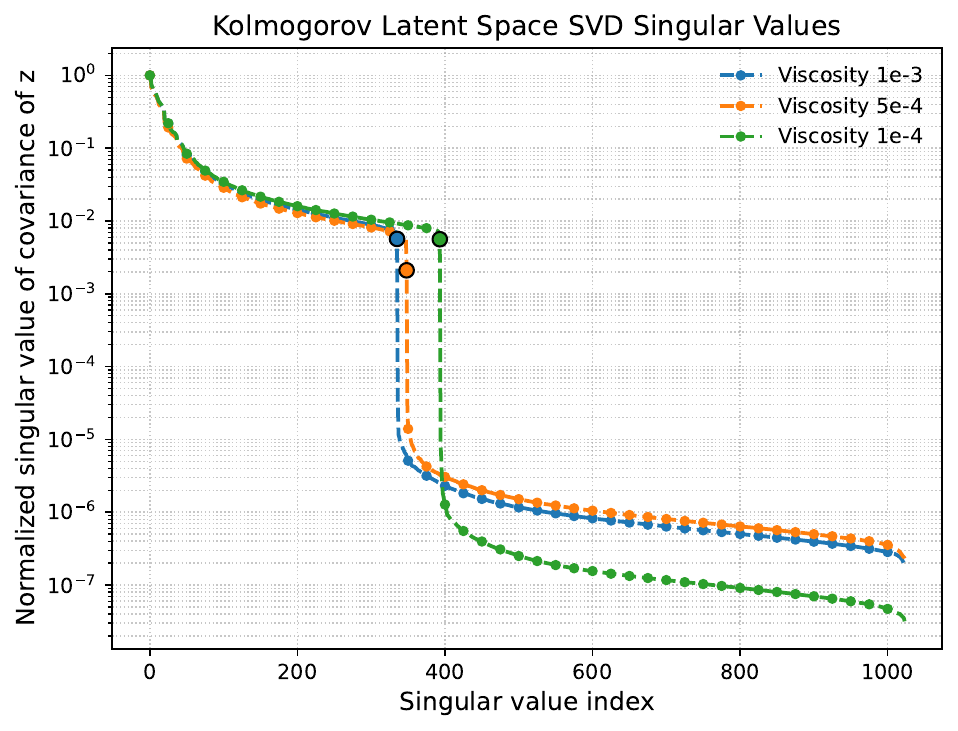}
  \caption{Ordered normalized singular values of SVD of latent space learned by the implicit rank minimizing autoencoder (Deep Compression Autoencoder) for Kolmogorov datasets. The cutoff dimensions are highlighted in the plot and have values 335, 348, and 393 respectively.}
  \label{fig:singular-val-dropoff-a}
\end{subfigure}
\hfill
\begin{subfigure}{.47\textwidth}
  \centering
  \includegraphics[width=\linewidth]{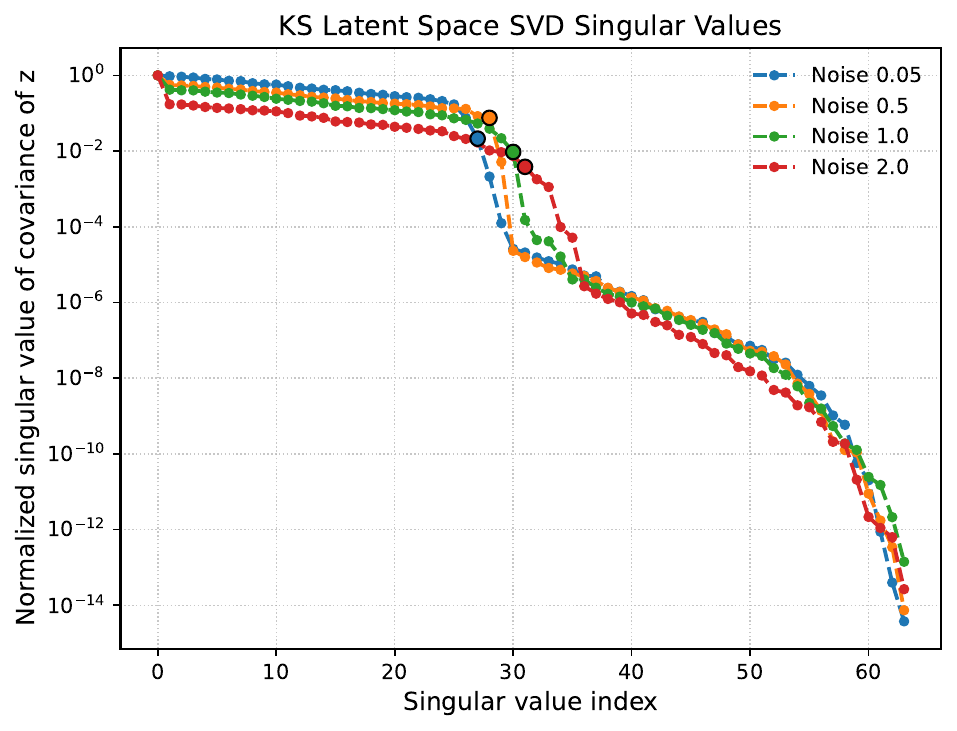}
  \caption{Ordered normalized singular values of SVD of latent space learned by the implicit rank minimizing autoencoder for Kuramoto-Sivashinsky $L=66$ datasets. The cutoff dimensions are highlighted in the plot and have values 27, 28, 30, and 31 respectively.}
  \label{fig:singular-val-dropoff-b}
\end{subfigure}
\caption{}
\label{fig:singular-val-dropoff}
\end{figure}

Classical reduced-order methods exploit low-dimensional structure in PDE dynamics \cite{fletcher1984computational}. Modern autoencoder approaches learn nonlinear latent coordinate transformations for systems evolving near attractors or manifolds \cite{strogatz2024nonlinear,koronaki2024nonlinear}. For stochastic systems, this perspective generalizes to random attractors \cite{crauel1997random}. We use implicit rank-minimizing autoencoders with weight decay \cite{jing2020implicit,zeng2024autoencoders} and SVD projection to discover low-rank latent spaces for accelerated stochastic interpolant generation. See Algorithm \ref{alg:dim-discovery} for details. We plot the singular values of the latent space for Kolmogorov datasets in Figure \ref{fig:singular-val-dropoff-a} and for KS datasets in Figure \ref{fig:singular-val-dropoff-b} with cutoff dimensions highlighted. As expected with implicit rank minimizing autoencoders, we see a characteristic dropoff in the singular values of the latent space which is especially clear in the Kolmogorov data.

\begin{algorithm}[h]
\small
\caption{Latent Dimension Discovery for Reduced Order Forecasting}
\label{alg:dim-discovery}
\begin{algorithmic}[1]
\REQUIRE One-step training pairs $\mathcal{D}=\{(x_i,x_i^+)\}_{i=1}^N$, maximum latent width $m$, weight decay $\lambda$
\ENSURE Encoder $E$, decoder $D$, projection basis $V_r$, reduced order model $G_r$

\STATE Initialize encoder $E_\theta:\mathbb{R}^d \rightarrow \mathbb{R}^m$, decoder $D_\phi:\mathbb{R}^m \rightarrow \mathbb{R}^d$, and stacked linear bottleneck $B_\psi:\mathbb{R}^m \rightarrow \mathbb{R}^m$.
\STATE Train the implicit rank-minimizing autoencoder with weight decay:
\[
(\theta,\psi,\phi)
\leftarrow
\arg\min_{\theta,\psi,\phi}
\frac{1}{N}\sum_{i=1}^N
\left\|
D_\phi\!\left(B_\psi(E_\theta(x_i))\right)-x_i
\right\|_2^2
+
\lambda
\left\|(\theta,\psi,\phi)\right\|_2^2 .
\]
\STATE Compute latent codes $z_i = B_\psi(E_\theta(x_i))$ and $z_i^+ = B_\psi(E_\theta(x_i^+))$.
\STATE Form the centered latent data matrix $Z = [z_1-\bar z,\ldots,z_N-\bar z]^\top$, where $\bar z = \frac{1}{N}\sum_{i=1}^N z_i$.
\STATE Compute the singular value decomposition $Z = U\Sigma V^\top$.
\STATE Choose the discovered latent dimension $r$ as the smallest rank whose singular values account for $99.99\%$ of the total singular-value mass:
\[
r =
\min \left\{
k :
\frac{\sum_{j=1}^{k} \sigma_j}
{\sum_{j=1}^{m} \sigma_j}
\geq 0.9999
\right\}.
\]
\STATE Let $V_r = [v_1,\ldots,v_r]$ be the leading $r$ right singular vectors.
\STATE Project one-step pairs to the discovered low-rank coordinates:
\[
y_i = V_r^\top(z_i-\bar z), \qquad
y_i^+ = V_r^\top(z_i^+-\bar z).
\]
\STATE Train a reduced order model $G_r$ on the reduced pairs $\{(y_i,y_i^+)\}_{i=1}^N$.

\STATE \textbf{Generation:} given an initial state $x_0$, compute
\[
y_0 = V_r^\top\left(B_\psi(E_\theta(x_0))-\bar z\right).
\]
\FOR{$n=0,\ldots,T-1$}
    \STATE Predict $y_{n+1} \sim G_r(\cdot \mid y_n)$
    \STATE Lift to the autoencoder latent space: $z_{n+1} = \bar z + V_r y_{n+1}$.
    \STATE Decode to the physical grid: $\hat{x}_{n+1} = D_\phi(z_{n+1})$.
\ENDFOR
\RETURN $E_\theta$, $B_\psi$, $D_\phi$, $V_r$, $G_r$
\end{algorithmic}
\end{algorithm}

\end{document}